\def\year{2021}\relax
\documentclass[letterpaper]{article} 
\usepackage{aaai21}  
\usepackage{times}  
\usepackage{helvet} 
\usepackage{courier}  
\usepackage[hyphens]{url}  
\usepackage{graphicx} 
\usepackage{natbib}  
\usepackage{caption} 
\usepackage{multirow}
\usepackage{xcolor}
\usepackage{bbding}
\usepackage{color}

\usepackage{amsmath}
\usepackage{lipsum}
\usepackage[pagebackref=true,breaklinks=true,colorlinks,bookmarks=false]{hyperref}

\usepackage[linesnumbered,ruled,vlined]{algorithm2e}
\usepackage{algpseudocode}  
\usepackage{amsmath}  

\title{Co-mining: Self-Supervised Learning for Sparsely Annotated Object Detection}
\author{
    Tiancai Wang\textsuperscript{\rm 1}, 
    Tong Yang\textsuperscript{\rm 1}, 
    Jiale Cao\textsuperscript{\rm 2},
    Xiangyu Zhang\textsuperscript{\rm 1}  
    \\
}
\affiliations{
    \textsuperscript{\rm 1} MEGVII Technology  \\
    \textsuperscript{\rm 2} Tianjin University \\
    \{wangtiancai, yangtong, zhangxiangyu\}@megvii.com, connor@tju.edu.cn

}

\begin{document}

\maketitle

\begin{abstract}
Object detectors usually achieve promising results with the supervision of complete instance annotations. However, their performance is far from satisfactory with sparse instance annotations. Most existing methods for sparsely annotated object detection either re-weight the loss of hard negative samples or convert the unlabeled instances into ignored regions to reduce the interference of false negatives. We argue that these strategies are insufficient since they can at most alleviate the negative effect caused by missing annotations. In this paper, we propose a simple but effective mechanism, called \textit{Co-mining}, for sparsely annotated object detection. In our \textit{Co-mining}, two branches of a Siamese network predict the pseudo-label sets for each other. To enhance multi-view learning and better mine unlabeled instances, the original image and corresponding augmented image are used as the inputs of two branches of the Siamese network, respectively. \textit{Co-mining} can serve as a general training mechanism applied to most of modern object detectors. Experiments are performed on MS COCO dataset with three different sparsely annotated settings using two typical frameworks: anchor-based detector RetinaNet and anchor-free detector FCOS. Experimental results show that our \textit{Co-mining} with RetinaNet achieves 1.4 \% $\sim$ 2.1\% improvements compared with different baselines and surpasses existing methods under the same sparsely annotated setting. Code is available at \url{https://github.com/megvii-research/Co-mining}.
\end{abstract}

\section{Introduction}

Object detection has achieved significant improvements in accuracy thanks to the development of deep Convolutional Neural Networks (CNNs) \cite{resnet16,Huang2017DenselyCC,VGG14,AlexNet12}. Most of object detectors are trained in the completely annotated object datasets \cite{coco14,Everingham_VOC_IJCV_2010,Object365}. However, when the object dataset is large in images and categories, it is pretty hard, even impossible, to annotate all existing object instances, especially in the crowded scenery. To deal with this problem, the task of Sparsely Annotated Object Detection (SAOD) is proposed recently, which has drew great attention
\cite{softsampling,samplingforsparse,brloss, SROD2020}. SAOD is different from conventional Semi-Supervised Object Detection (SSOD).
In SSOD \cite{CSD2019,Tang2020proposal,NOTERCNN}, a small part of training images are fully annotated and the rest of training images are totally unlabeled. Compared to SSOD that annotates a part of images, SAOD annotates a part of instances in each image. 

\begin{figure}[t]
  \centering
  \includegraphics[width=0.45\textwidth]{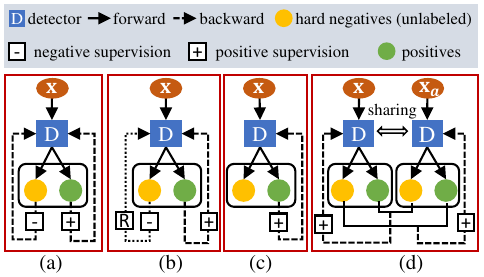}\vspace{-0.3cm}
  \caption{Comparison of different pipelines for sparsely annotated object detection (SAOD). For simplicity, we only show positives and hard negatives. Here, hard negatives mainly represent unlabeled instances.  (a) The standard object detector directly used for SAOD. (b) Soft sampling strategy and background recalibration loss that re-weights hard negatives. (c) Part-aware sampling strategy that ignores the loss of part categories for some hard negatives. (d) Instead of reweighting and ignoring the loss of hard negatives, our \textit{Co-mining} exploits to convert the hard negatives for positive supervision by using a Siamese network with two inputs.}
\label{fig:archcom}
\end{figure}

One of major challenges in SAOD is that the unlabelled instances will interference the training of a detector. During training, the unlabelled instances provide an incorrect supervision signal, where those unlabeled instances and their surrounding proposals are mistakenly regarded as hard negatives in Fig. \ref{fig:archcom}(a). As a result, the weight updated in the gradient back-propagation will be misguided. Inspired by the practice in semi-supervised object detection, one straightforward solution for SAOD \cite{samplingforsparse} is to first train a detector with sparse annotations and then use the pseudo labels generated by the learned model to retrain a new detector. However, since the first learned model is deeply confused by the unlabeled instances, the generated pseudo labels for the second detector learning are still of low credibility. Hence, how to deal with the unlabelled instances during training is a key factor to improve SAOD.


Recently some effective sampling strategies \cite{softsampling,samplingforsparse,brloss, sun2020mining} are proposed to tackle the problem of sparse annotations. Soft sampling \cite{softsampling} in Fig. \ref{fig:archcom}(b) re-weights the loss for the RoIs based on their overlaps with the annotated instances. 
Background recalibration loss (BRL) proposed in \cite{brloss} regards the unlabeled instances as hard-negative samples and re-calibrates their losses.
Part-aware sampling \cite{samplingforsparse} in Fig. \ref{fig:archcom}(c) ignores the classification loss for part categories whose instance is the component of subject categories. Despite these recent advances, the performance in SAOD is still far from satisfactory compared to the completely annotated counterparts.  Essentially, these approaches aim to reduce the negative effect of incorrect supervision by  using the re-weighting or ignore strategies.  As a result, these strategies can at most alleviate the confusing supervision and fail to make full use of the useful information mined from unlabeled instances. Therefore,  a natural question to ask is: why not convert the category of unlabeled instance into the positive category to help improve detection performance during training?

In this paper, we explore a simple but effective mechanism, called \textit{Co-mining}, for sparsely annotated object detection.  Our \textit{Co-mining}  aims to mine positive supervision information from the  unlabeled instances, which can be built on both the anchor-based and anchor-free detectors.
For both anchor-based and anchor-free detectors, the network can be simply divided into two parts: the backbone and the detection head. The backbone part \cite{resnet16,VGG14} mainly contains a deep CNN model pretrained on the large-scale image classification dataset, like ImageNet \cite{ImageNet}. To better deal with scale-variance problem in object detection, the deep CNN model is usually equipped with feature pyramid network (FPN) \cite{Tsung-YiLinCVPR17}, image pyramid \cite{EFIP2019}, or other architectures \cite{wang2020HRNET}. Based on the features generated from the backbone part, the detection head part is employed to generate the final predictions, including classification score and regression offset. To effectively mine the unlabeled instances and convert them into foreground classes, our \textit{Co-mining} in Fig. \ref{fig:archcom}(d) proposes to employ a Siamese network where two branches share the weights. During training, these two branches generate the pseudo-label sets online by a co-generation module. The pseudo-label set generated from one branch will merge with the sparsely annotated-label set, resulting in a more complete set used as the positive supervision signal of another branch. Inspired by the recent self-supervised learning \cite{simCLR,moco,byol} and novelty detection approach \cite{choi2020novelty}, the original image and corresponding augmented image are used as two different inputs of Siamese network. With different input images, two branches can learn different and multi-view feature patterns, which can improve pseudo-label learning.

Experiments are conducted on the sparsely annotated versions of MS COCO dataset \cite{coco14}. We follow the settings in \cite{brloss} to generate corresponding datasets for four cases (full, easy, hard, and extreme) by using different instance drop-rates. We perform a thorough ablation study to demonstrate the effectiveness of our approach. Experimental results show that our proposed \textit{Co-mining} with RetinaNet achieves 1.4\% $\sim$ 2.1\% improvements under partial-label conditions. We also validate the effectiveness of our \textit{Co-mining} on
anchor-free detector FCOS \cite{FCOS}. Moreover, the RetinaNet detector with our \textit{Co-mining} can achieve superior performance compared with the state-of-the-art methods. 

\section{Related Work}


\noindent \textbf{Completely annotated object detection:} Recent years have witnessed great progress in
object detection thanks to the development of CNN. Generally, CNN-based object detection can be divided into two groups: anchor-based and anchor-free approaches. Anchor-based approaches can be further divided into two-stage and single-stage approaches. Two-stage approaches \cite{fasterrcnn_2015_nips,rfcn16,Cao2019D2Det,Wu2020Forest} firstly generate some possible proposals for objects and then classify/regress these proposals by a RoI head network. Different from two-stage approaches, single-stage approaches \cite{WLiu16ECCV,YOLOv1,wang2019lrfnet,Nie_2019_ICCV} directly predict the classes and the offsets for pre-defined anchors. Recently, anchor-free approaches \cite{FCOS,Cornernet,Duan&CenterNet} have been proposed to eliminate the requirement of setting pre-defined anchors. For example, CornerNet \cite{Cornernet} and CenterNet \cite{Zhou&CenterNet} treat object detection as keypoint detection problem, while FCOS \cite{FCOS} densely predicts the left, top, right, and bottom distance to the object boundary at each foreground position. Usually, both the anchor-based approaches and the anchor-free approaches are trained on completely annotated object datasets.

\noindent \textbf{Sparsely annotated object detection (SAOD):}
Different from generic object detection having complete annotations, SAOD faces a part of unlabeled instances in each image. These unlabeled instances give an incorrect supervision signal, which can confuse the detector learning. To solve this problem, existing methods either re-scale or ignore the loss of hard negative samples.
To balance the unlabeled instances and background in background regions, soft sampling \cite{softsampling} assigns a low weight to the hard negative RoIs. Similar to soft sampling, background recalibration loss \cite{brloss} re-calibrates the loss of negative when the negative has a high positive confidence, which is designed for single-stage detector. Inspired by the human intuition for hierarchical relation between different objects, part-aware sampling \cite{samplingforsparse} ignores the loss of some part categories whose instances are inside the instances with subject categories.


\begin{figure*}[t]
  \centering
  \includegraphics[width=1.0\textwidth]{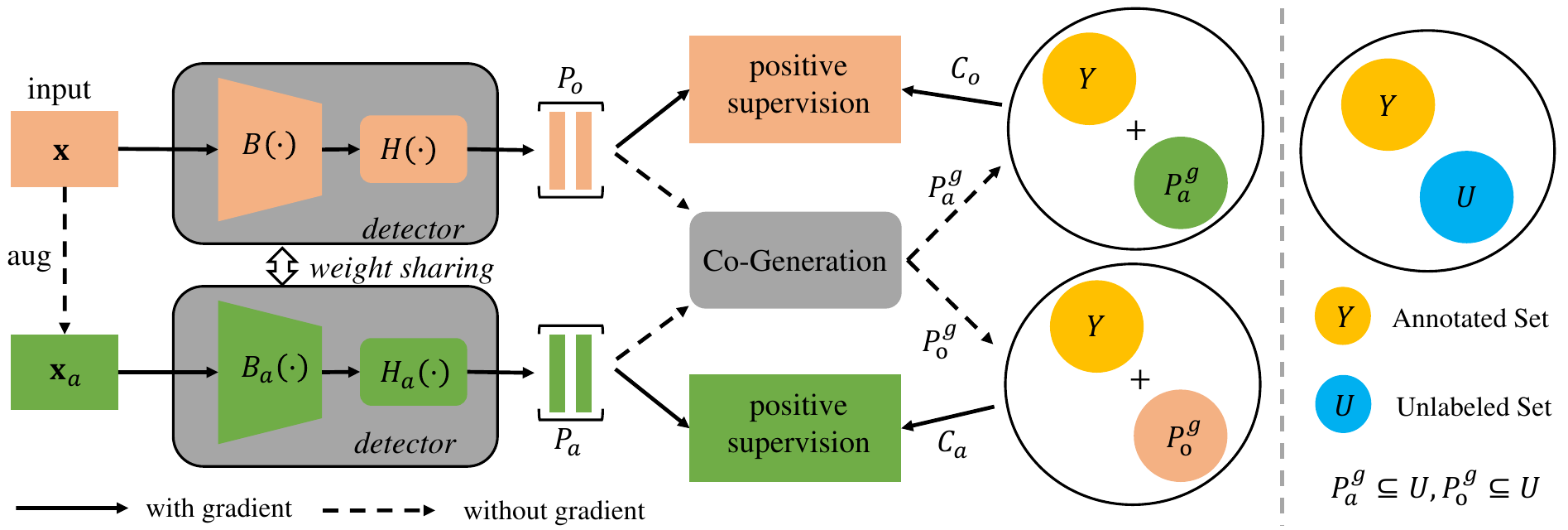}\vspace{-0.2cm}
  \caption{The overall architecture of the proposed \textit{Co-mining} mechanism for sparsely annotated object detection. The input image ${\bf x}$ and corresponding augmented image ${\bf x}_a$ are two different inputs to the Siamese network that contains the backbone $B$ and prediction head  $H$. The two branches in the Siamese network shares the parameters. With two different inputs, two branches has two different predictions $P_{o}$ and $P_{a}$, respectively. After that, $P_{o}$ and $P_{a}$ go through the co-generation module to generate the pseudo-label sets $P^g_{a}$ and $P^g_{o}$. The pseudo-label set generated from one branch will merge with annotated ground-truth set $Y$ to generate the complementary ground-truth set $C$, which is employed as the positive supervision signal for another branch.}
\label{fig:arch}
\end{figure*}

\noindent \textbf{Semi-supervised object detection:} To reduce the expensive costs of annotating instances in a large number of images, semi-supervised object detection \cite{CSD2019,NOTERCNN,Tang2020proposal,wang2018towards,zhao2019sess} focuses on  leveraging a small part of labeled images and a large part of unlabeled/weakly labeled images to improve  performance. CSD \cite{CSD2019} applies consistency constraint on the predictions of the original and flipped images to make full use of available unlabeled data. Furthermore, SESS \cite{zhao2019sess} utilizes consistency losses to enforce the consistency between two sets of predicted 3D object proposals to mitigate the issue of expensive annotation. In addition, NOTE-RCNN \cite{NOTERCNN} proposes a new iterative training-mining pipeline to mine bounding boxes from the unlabeled data. Different from the semi-supervised detection methods, our method focuses on sparsely annotated object detection. 


\noindent \textbf{Co-training Learning:} It aims to use two classifiers to handle unlabeled data \cite{cotrain} or noisy data \cite{coteach,yu2019does,JoCoR}. \cite{cotrain} utilizes two distinct views of samples to predict pseudo labels for unlabeled data and expands labeled data to solve semi-supervised problem. As two classifiers (networks) have different learning abilities, Co-teaching \cite{coteach} uses these classifiers to filter different noisy data and exchanges clean data of two classifiers for updating their own parameters. Unlike these previous works, our proposed \textit{Co-mining} is designed for sparsely annotated object detection, where part of instances are unlabeled in each training image. Meanwhile, compared to these works having two distinct networks, our \textit{Co-mining} adopts a Siamese network strategy that has two branch during training and one branch during inference. Thus, our \textit{Co-mining} does not add any extra computational costs during inference.

\section{Our Co-mining}

Our goal is to mine the positive supervision from the unlabeled instances to help improve sparsely annotated object detection. To achieve this goal, a \textit{Co-mining} mechanism with a Siamese network is proposed.  In this section, we will first clarify the motivation of our proposed \textit{Co-mining} and then describe the process of our approach in detail.

\subsection{Motivation}
As discussed earlier, existing methods either re-scale the loss of hard negatives \cite{brloss,softsampling,yang2020puleaning} or set the part categories of unlabeled regions into ignore class \cite{samplingforsparse}. We argue that both the re-weighting strategy and ignore strategy are sub-optimal since these strategies at most alleviate the negative effect of incorrect gradient caused by unlabeled instances and fail to make full use of the information from unlabeled instances. Thus, a natural question to ask is: why not mine the unlabeled instances and convert them as the positive supervision signal for improving detection performance during training?

Motivated by the advances in the co-training mechanism \cite{cotrain,coteach,JoCoR} and the self-supervised learning \cite{moco,simCLR,ye2019cvpr}, we develop a simple but effective mechanism, called \textit{Co-mining}, for sparsely annotated object detection. We propose to construct a Siamese detection network during training. The Siamese detection network  has two branches that generate two different pseudo-label sets using the proposed co-generation module.  The pseudo-label set generated from one branch will merge with the sparsely annotated-label set, resulting in a more complete label set. This complete label set will be used as the supervision signal of another branch. To generate the correlated multi-view data and learn different feature patterns, the original images and corresponding augmented images are used  as two different inputs of the Siamese detection network.

\subsection{Overall Architecture}

The overall architecture of our \textit{Co-mining} is shown in Fig.~\ref{fig:arch}. Given a training image ${\bf x}$, $Y$ represents the set of sparsely annotated instances while $U$ denotes the set of unlabeled instances. The Siamese network is adopted for feature extraction and detection prediction. Each branch of the Siamese network contains a backbone $B$ and prediction head $H$. The pseudo-labels generated by the proposed co-generation module  are used to supervise model learning during training. The detailed pipeline is introduced as follows.


The input $\bf x$ goes through the top branch ($B$ and $H$) of the Siamese network to generate the prediction $P_o$. Meanwhile, the corresponding  augmented image ${\bf x}_{a}$ goes through the bottom branch ($B_a$ and $H_a$) of the Siamese network to generate the prediction $P_a$. The prediction from each branch contains the outputs of classification and regression. For the anchor-free detector FCOS, the prediction also contains the centerness output. For simplicity, we ignore centerness output. Then the predictions $P_o$ and $P_a$ both go through the proposed co-generation module and generate two pseudo-label sets $P^g_{o}$ and $P^g_{a}$ corresponding to the unlabeled instance set $U$.  The generated pseudo-label set $P^g_{o}$ merge with the annotated ground-truth set $Y$ to generate the more complete set $C_{a}$. $C_{a}$ is then used to guide the training of the head $H_{a}$ and the backbone $B_a$ in bottom branch. Similarly, the more complete set $C_{o}$ combining $P^g_a$ and $Y$ is used to guide the training of $H$ and $B$ in top branch. Note that, the solid arrows in Fig.~\ref{fig:arch} do not have  the gradient back-propagation. For more details, please refer to Algorithm~\ref{comine_algorithm}.

\subsection{Data Augmentation}






Data augmentation has been widely used in fully/semi-supervised learning, which can make the detector more robust to the variances in sizes, shapes, and appearances. Recently, data augmentation is introduced in contrastive learning \cite{simCLR} or out-of-distribution detection \cite{choi2020novelty}. In our \textit{Co-mining}, some widely used data augmentation strategies, such as Gaussian blur and color-jitter, can be employed to obtain the augmented image ${\bf x}_{a}$. 





\noindent \textbf{Why using data augmentation in Co-mining?}
Different from the objective of data augmentation in self-supervised learning \cite{simCLR,moco}, which aims to maximize the data agreement after feature projection, data augmentation in our \textit{Co-mining} targets to generate two rich but diverse pseudo-label sets for improving object detection. The two  pseudo-label sets are used as the positive supervision to guide the training of two branches, respectively.
With the help of these diverse pseudo-labels, the two branches help each other to avoid the incorrect interference of unlabeled instances as much as possible. Some quantitative and qualitative experiments are given in experiment section.

\begin{algorithm}[t]
  \caption{Our \textit{Co-mining} Algorithm}  
  \label{comine_algorithm}
  \KwIn{Siamese network $f_{det}$ with the  parameters ${\bf w} = \{{\bf w}_{b},{\bf w}_{h}\}$, the number of max iterations $N_{max}$, detection score threshold $\tau$, $IoU$ thresholds $\theta_{1},\theta_{2}$, learning rate $\eta$.}
  Shuffle the training set $D$\;
  \For{$k=1,2,...,N_{max}$}  
  {  
   Fetch mini-batch $D_{k}$ from $D$\;
   \textbf{Augment} ${\bf x}$ to obtain ${\bf x}_{a}$, $\forall$ ${\bf x}$ $\in$ $D_{k}$\;
   $P_{o}$ = $f_{det}({\bf x}, {\bf w}_{b}, {\bf w}_{h})$,
   $P_{a}$ = $f_{det}({\bf x}_a, {\bf w}_{b}, {\bf w}_{h})$\;
  \textbf{Detach} $P_{o}$, $P_{a}$ and convert them to initial pseudo label sets $P^g_{o}$, $P^g_{a}$\;
    \For{$box_{i}$ \text{in} $P^g_{o}$ or $P^g_{a}$} 
    {  
      \If{$b_{i}^{score}<\tau$}  
      {  
        Remove $box_{i}$ from $P^g_{o}$ or $P^g_{a}$\;  
      }
      }
    Perform NMS on $P^g_{o}$ or $P^g_{a}$ with threshold $\theta_{1}$;\\
    \For{$box_{i}$ \text{in} $P^g_{o}$ or $P^g_{a}$} 
    {
      \textbf{Compute} the $IoU$ matrix with annotated ground-truth set $Y$\;
      \If{$IoU(box_{i},box_{gt})>\theta_{2}$, $\forall$ $box_{gt}$ $\in$ $Y$}  
      {  
        Drop $box_{i}$ from $P^g_{o}$ or $P^g_{a}$\;  
      }
    }
    \textbf{Calculate} the overall loss $L$ by Eq. \ref{overall_loss};\\
    \textbf{Update} the weights ${\bf w}_b$ and ${\bf w}_h$ in backbone and prediction head.
  }  
  \KwOut{${\bf w} = \{{\bf w}_b,{\bf w}_h\}$} 
\end{algorithm}

\subsection{Co-Generation}

As mentioned in Section Overall Architecture, the proposed co-generation module is employed to convert the predictions of one branch into the pseudo-labels for another branch. In each branch, the predictions contains classification confidence and regression offset. Based on classification confidence and regression offset, the densely predicted bounding boxes with the classification scores can be decoded. Using the dense predictions from two branches, we can generate the pseudo-labels for each other as the following steps:

\textbf{Step1: Filter the bounding boxes with lower scores.} Usually, the predicted bounding box with a lower confidence score has a little chance of detecting unlabeled instances. Thus, the dense predicted results can be first filtered by a given score threshold.

\textbf{Step2: Remove redundant boxes by Non-Maximum Suppression (NMS).} 
Considering the objects are densely detected,
one instance will be detected by many surrounding bounding boxes. As a result, the bounding boxes detected around one instance will highly overlap. In this paper, we simply keep the bounding box with highest confidence score to represent each instance by NMS.


\textbf{Step3: Drop the bounding boxes representing annotated instances.}
The kept bounding boxes contain both unlabeled instances and annotated instances. To avoid generating duplicated labels for the annotated instances, we remove the bounding boxes representing the annotated instances according to their overlaps with annotated instances.

Based on the outputs of two branches, the proposed co-generation module can generate two pseudo-label sets $P^g_{o}$ and $P^g_{a}$, respectively. $P^g_{o}$ and $P^g_{a}$ will merge with the annotated ground-truth set $Y$ to generate two more complete sets $C_{a}$ and $C_{o}$, respectively. Note that the gradient back-propagation of proposed co-generation module is disabled during training.

\subsection{Complementary Supervision}

In our \textit{Co-mining}, two branches of Siamese network generates complementary pseudo-labels for each other using the proposed co-generation module. The pseudo-label set $P^g_o$ generated by top branch merges with the sparsely annotated ground-truth set $Y$, which provide one more complete instance set $C_{a}$. Similarly, the more complete instance set $C_{o}$ is generated by $P^g_a$ and $Y$. Namely, the more complete instance sets $C_{o}$ and  $C_{a}$ can be represented as follows:
\begin{equation}
\centering
\begin{aligned}
C_o&=P^g_a\cup Y,\\
C_a&=P^g_o\cup Y.
\end{aligned}
\end{equation}
The sets $C_{o}$ and $C_{a}$ are used as the supervisions of these two branches.
In this way, the overall loss for complementary supervision can be written as
\begin{equation}
\label{overall_loss}
\begin{aligned}
L&=L_{o}+L_{a}\\
 &=L^{c}(P_{o},C_{o})+L^{r}(P_{o},C_{o})+L^{c}(P_{a},C_{a})+L^{r}(P_{a},C_{a})\\
 &=L^{c}(P_{o}({\bf x},{\bf w}_{b},{\bf w}_{h}),C_{o})+L^{r}(P_{o}({\bf x},{\bf w}_{b},{\bf w}_{h}),C_{o})\\
 &+L^{c}(P_{a}({\bf x}_{a},{\bf w}_{b},{\bf w}_{h}),C_{a})+L^{r}(P_{a}({\bf x}_{a},{\bf w}_{b},{\bf w}_{h}),C_{a}) 
\end{aligned}
\end{equation}
where $L_{o}$ and $L_{a}$ are the losses for top branch and bottom branch in Siamese network, respectively. ${\bf w}_{b}$ and ${\bf w}_{h}$ represent the weights of backbone and prediction head, which are shared by two branches of Siamese network. $L^{c}$ and $L^{r}$ are the loss function for classification and regression. For classification task, the $L^{c}$ is usually the focal loss defined in \cite{RetinaNet}. For the regression task, the $L^{r}$ is different in RetinaNet and FCOS. FCOS adopts IoU loss while RetinaNet employs smooth $L_{1}$ loss.

\section{Experimental Settings}
\noindent \textbf{Dataset:} All experiments are conducted on the challenging MS COCO dataset \cite{coco14}. 
The COCO-2017 train set with sparse annotations is used for training.
Following the way in \cite{brloss}, some annotations in training set are randomly erased to generate three training sets: easy, hard and extreme sets. For a given image with full annotations, the \textbf{easy} set randomly erases one instance annotation, the \textbf{hard} set randomly erases half of annotations, while the \textbf{extreme} set only preserves one annotation. Further, to fairly compare with state-of-the-art methods in literature, experiments are also conducted under the \textbf{COCO-50miss} \cite{brloss} training set where 50\% annotations per object category are randomly erased. The COCO-2017 validation set with complete annotations is used for all performance evaluations.
The standard COCO-style Average Precision (AP) is used as evaluation metric.

\noindent \textbf{Detector:} Experiments are performed on two typical detectors: anchor-based RetinaNet \cite{RetinaNet} and anchor-free FCOS \cite{FCOS}. The backbone model ResNet-50 \cite{resnet16} is used for ablation study and the deep model ResNet-101 \cite{resnet16} is used for comparison with other state-of-the-art methods. We adopt 8 TITAN 2080ti GPUs with a batch size of 16 for training. During training, there are 90$k$ iterations in total. The learning rate is initially set to 0.01 and gradually decreases to 0.001 and 0.0001 at 60$k$ and 80$k$ iterations. Warm-up strategy adopted for the first 1$k$ iterations to stabilize the training process.



\section{Experimental Results}

\subsection{Ablation Study}

\begin{table}[t]
  \centering
  \begin{tabular}{l|cccc}
    \hline
    Set &  Original     & Augmented  & Joint  & Co-mining\\
    \hline
    Full & 36.0 & 35.4 & 36.4 & \textbf{36.8}\\
    Easy & 33.4 & 32.9 & 33.8 & \textbf{35.4}\\
    Hard &  30.4 & 30.8 & 30.6 & \textbf{31.8} \\
    Extreme & 20.9 & 21.4 & 21.2 & \textbf{23.0} \\
    \hline
  \end{tabular}\vspace{-0.3cm}
  \caption{Comparison of our \textit{Co-mining} and three different baselines trained under four different annotation sets. The anchor-based RetinaNet is adopted as the detector.  `Original' means that the network is trained without augmentations and `Joint' means that the network is trained with both original and augmented inputs. Our \textit{Co-mining} outperforms three baselines under all the sets, even the full annotated set.}
  \label{tbl:retinanet}
\end{table}

\begin{figure*}[t]
  \centering
  \includegraphics[width=1.0\textwidth]{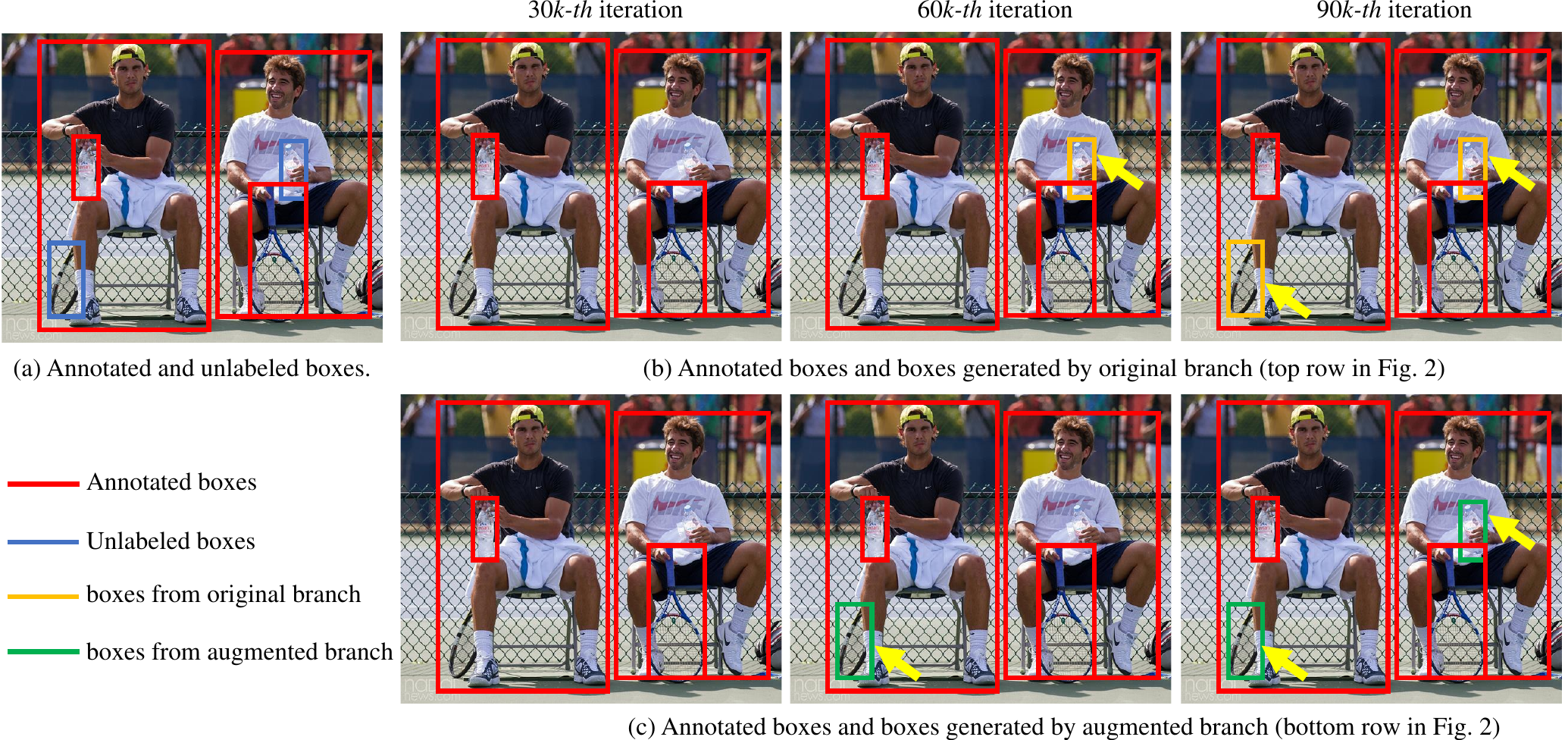}\vspace{-0.1cm}
  \caption{Visualization of the pseudo labels (bounding boxes) generated by two different branches of Siamese network at different iterations of the training.}
\label{fig:vislabel}
\end{figure*}

Our \textit{Co-mining} uses a Siamese network with the inputs of both original and augmented images. To demonstrate that our improvement does not simply come from data augmentation and joint training, we compare our \textit{Co-mining} with three baseline training strategies (named as `original', `augmented', and `joint') under four different training sets in Tab.~\ref{tbl:retinanet}. `original' and `augmented' use the original images and the augmented images as the inputs of detector, respectively. `joint' adopts the same Siamese structure of our \textit{Co-mining} without co-generation module. Different from our \textit{Co-mining}, `original', `augmented', and `joint' does not mine unlabeled instances as positive supervisions.  
Our \textit{Co-mining} outperforms these three baseline strategies under all the four sets. For example, under the extreme set, our \textit{Co-mining} is 2.1\% better than `original',  1.6\% better than `augmented', and  1.8\% better than `joint'. The reason that our proposed method outperforms these baselines can be explained as follows: our \textit{Co-mining} strategy can mine these unlabeled instances to reduce the incorrect signal interference and use them as the positive supervisions to help guide the detector learning. 
Note that our \textit{Co-mining} outperforms `Joint'  even on the full annotation set. It demonstrates that original full annotations still have some unlabeled instances and our \textit{Co-mining} can mine these unlabeled instances.
\begin{table}[t]
  \centering
  \resizebox{\linewidth}{!}{
  \begin{tabular}{cc|ccc}
    \hline
    co-generation &Siamese network &$AP$ &$AP_{50}$ &$AP_{75}$ \\
    \hline
     &    &30.4   &49.4  & 32.0 \\
     \Checkmark  &  & 30.9   & 49.3  &32.6  \\
        &\Checkmark  & 30.6   & 49.3  &32.5  \\
      \Checkmark  & \Checkmark & \textbf{31.8}   &\textbf{50.3}  &\textbf{33.7}  \\
    \hline
  \end{tabular}
 }\vspace{-0.3cm}
 \caption{Impact of integrating different modules in our \textit{Co-mining}  trained on the hard set.}
 \label{tbl:multiple_head}
\end{table}

\begin{table}[t]
  \centering
  \begin{tabular}{l|ccccc}
    \hline
    Threshold & 0.4 & 0.5 & 0.6 & 0.7 & 0.8\\
    \hline
    $AP$    & Nan   & 30.6  & \textbf{31.8}  & 30.7 & 30.7 \\
    \hline
  \end{tabular}\vspace{-0.3cm}
  \caption{Impact of score thresholds trained on the hard set.}
  \label{tbl:score_thres}
\end{table}

\begin{table}[t]
  \centering
  \begin{tabular}{l|cccccc}
    \hline
    Method &$AP$ &$AP_{50}$ &$AP_{75}$ &$AP_{s}$ &$AP_{m}$ &$AP_{l}$\\
    \hline
    baseline & 30.4   & 49.4  & 32.0  & 15.8 & 32.6 & 41.2\\
    w/o aug    & 30.6   & 48.4  & 32.4  & 15.5 & 32.4 & 41.7\\
    blur    & 31.2   & 49.4  & 33.2  & 15.7 & 33.4 & 42.4\\
    color & \textbf{31.8}   & \textbf{50.3}  & \textbf{33.7}  & \textbf{16.2} & \textbf{34.0} & \textbf{43.3}\\
    \hline
  \end{tabular}\vspace{-0.3cm}
  \caption{Impact of different data augmentation strategies.}\vspace{-0.2cm}
  \label{tbl:data_aug}
\end{table}

Our \textit{Co-mining} contains two modules: co-generation module and Siamese network module. Co-generation module mines and converts the unlabeled instances, and Siamese network module has two branches with two inputs of the original and augmented images. Tab.~\ref{tbl:multiple_head} shows the impact of integrating different modules. Integrating our modules to the baseline can gradually improve detection performance. When only adding co-generation module to the baseline, a single detector with the input of original data is supervised by pseudo labels generated by itself. The improvement of co-generation module is 0.5\%. With both co-generation module and Siamese network, the improvement is 1.4\%. It demonstrates that Siamese network with two different inputs generate more diverse pseudo labels, which can further improve detection performance. To better verify it, Fig. \ref{fig:vislabel} visualizes the results of pseudo labels generated by two branches of Siamese network at different iterations. At the 60$k$-$th$ iteration, the original branch (top branch in Fig. \ref{fig:vislabel}) generates the pseudo label of unlabeled bottle, and the augmented branch  (bottom branch in Fig. \ref{fig:vislabel}) generates the pseudo label of unlabeled tennis racket. With the supervision of mined instances, two branches of Siamese network both detect unlabeled instances at the 90$k$-$th$ iteration.

\begin{table}[t]
  \centering
  \begin{tabular}{l|cccc}
    \hline
    Set & Original     & Augmented  & Joint  & Co-mining\\
    \hline
    Full & 38.4 & 37.8 & 39.1 & \textbf{39.3}\\
    Easy & 36.3 & 35.7 & 36.8 & \textbf{37.6}\\
    Hard & 33.2 & 33.5 & 33.5 & \textbf{34.4} \\
    Extreme & 22.1 & 22.3 & 22.8 & \textbf{23.1} \\
    \hline
  \end{tabular}\vspace{-0.2cm}
  \caption{Comparison of our \textit{Co-mining} and different baselines using anchor-free FCOS.}\vspace{-0.2cm}
  \label{tbl:fcos}
\end{table}

\begin{figure*}[t]
  \centering
  \includegraphics[width=1.0\textwidth]{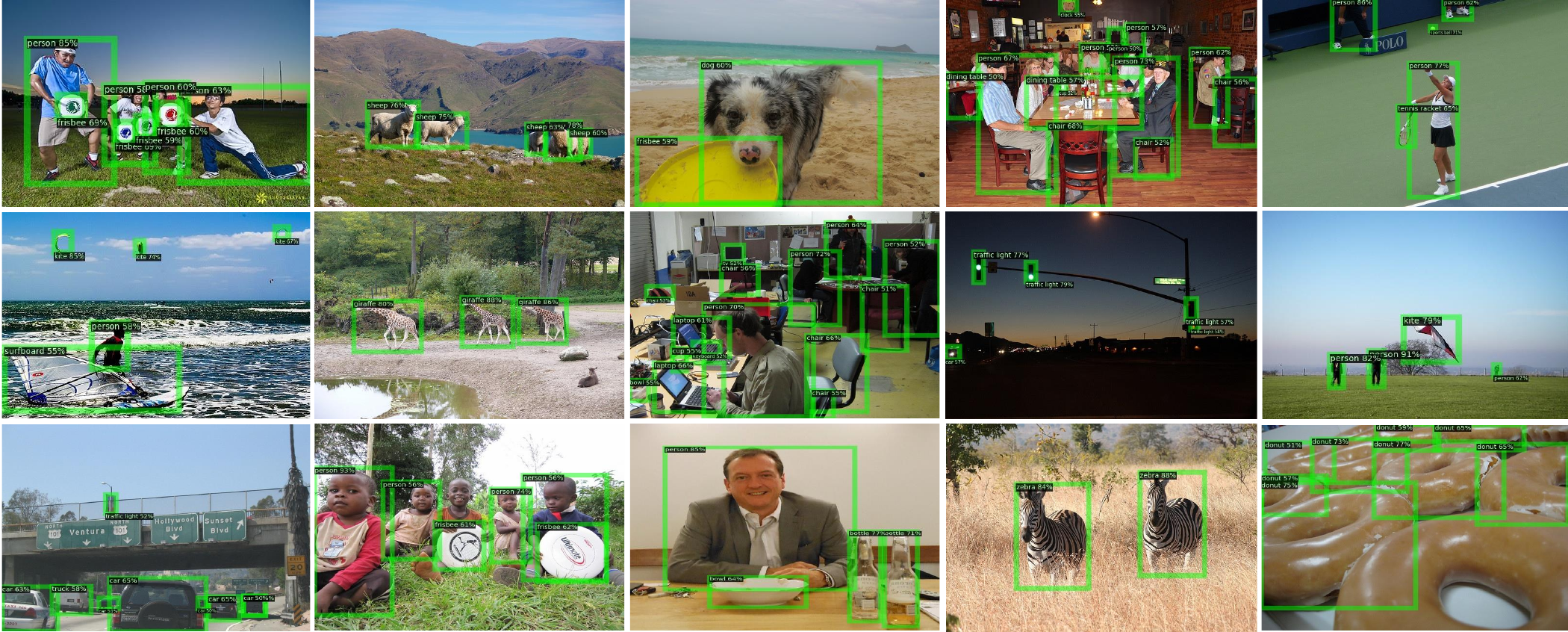}\vspace{-0.3cm}
  \caption{Some qualitative results of our \textit{Co-mininig} with the anchor-based RetinaNet on the COCO-2017 validation set.}
\label{fig:visualize}
\end{figure*}

\begin{figure*}[t]
  \centering
  \includegraphics[width=\textwidth]{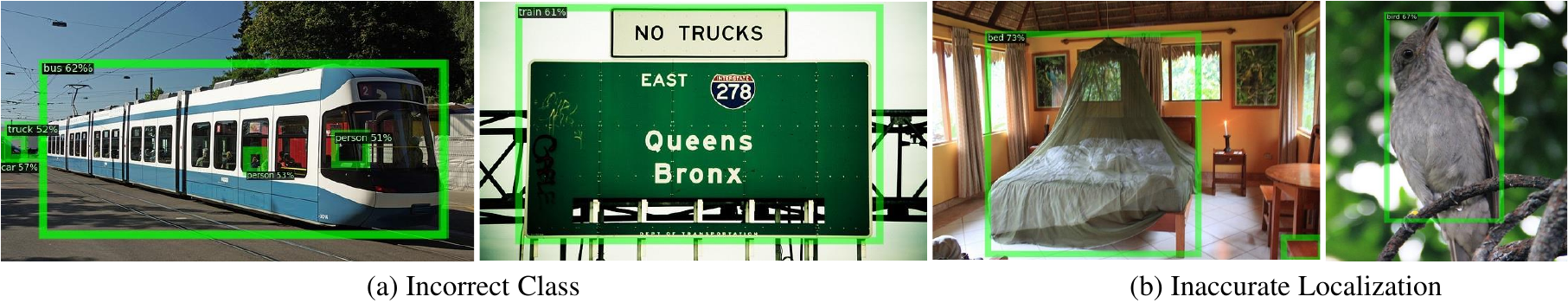}\vspace{-0.3cm}
  \caption{Some failure predicted pseudo labels of our proposed \textit{Co-mining} during training, including incorrect classification and inaccurate localization.}
\label{fig:failure_cases}
\end{figure*}

In our co-generation module, the score threshold $\tau$ determines the number of pseudo-labels used as positive supervision. Tab.~\ref{tbl:score_thres} shows the impact of different score thresholds trained on the hard set. Co-generation with the threshold of 0.6 produces the best performance. Using the lower threshold of 0.4 leads to loss explosion because many incorrect pseudo labels are used for supervision. On the contrary, using the higher thresholds (\textit{e.g.,} 0.7 and 0.8) may lose many unlabeled instances, which also results in a lower performance. In our Siamese network module, data augmentation is used for one branch. Tab.~\ref{tbl:data_aug} compares different data augmentation strategies, including no augmentation, blur augmentation, and color augmentation. Color augmentation provides the best performance. 

\subsection{Generalization on Anchor-free Detectors}
Compared to anchor-based methods, anchor-free methods \cite{FCOS,Cornernet,Duan&CenterNet} avoid parameter settings of anchors. To show the generalization of our \textit{Co-mining},
Tab.~\ref{tbl:fcos} compares our method with three baselines under four different sets using anchor-free FCOS. FCOS with our \textit{Co-mining} outperforms three baselines on all the four sets. It demonstrates that our \textit{Co-mining} is suitable to various modern detectors. In the future, we will exploit the generalization on two-stage object detectors.

\subsection{Comparison with Other Methods}

Tab.~\ref{tbl:compare_sota} compares our method using RetinaNet with state-of-the-art methods trained on COCO-50miss set and tested on COCO-2017 validation set. For fair comparison, all the methods use ResNet-101 as the backbone. Because pseudo label method  and part-aware method in \cite{samplingforsparse} do not provide source code, we re-implement them on RetinaNet. Our \textit{Co-mining} outperforms BRL \cite{brloss} and part-aware sampling \cite{samplingforsparse} by 1.2\% and 5.5\%. Note that, our \textit{Co-mining} achieves again of 2.2\% on $AP_{50}$ compared to BRL. It indicates that our \textit{Co-mining} improves classification ability of the detector by mining unlabeled instances. Fig.~\ref{fig:visualize} shows some qualitative results.

\begin{table}[t]
  \centering
  \resizebox{\linewidth}{!}{
  \begin{tabular}{l|cccccc}
    \hline
    Method &$AP$ &$AP_{50}$ &$AP_{75}$ &$AP_{s}$ &$AP_{m}$ &$AP_{l}$\\
    \hline
    baseline\cite{brloss}    & 23.7   & 40.6  & 33.5  & - & - & -\\
    Pseudo label\cite{samplingforsparse}    & 27.5   & 46.3  & 34.5  & 15.1 & 33.2 & 41.7\\
    Part-aware\cite{samplingforsparse}
    & 28.4   & 46.9  & 35.1  & 15.3 & 33.6 & 42.0\\
    BRL\cite{brloss} & 32.7   & 50.8  & 35.3  & - & - & -\\
    Ours & \textbf{33.9}   & \textbf{53.0}  & \textbf{36.1}  & \textbf{18.7} & \textbf{36.8} & \textbf{45.7}\\
    \hline
  \end{tabular}
 }\vspace{-0.2cm}
 \caption{Comparison with state-of-the-art methods trained on the COCO-50miss set with the backbone ResNet101.}
  \label{tbl:compare_sota}
\end{table}

\subsection{Discussions} 
\noindent \textbf{Limitations:} Though our \textit{Co-mining} can mine most of unlabeled instances, it sometimes provides some failure pseudo labels for classification or regression. Fig. \ref{fig:failure_cases} shows some failure examples of wrong class or inaccurate localization. 

\noindent \textbf{Impact of false positives:} At the beginning training stage, the detector employed is not fully trained, whose predicted classification scores tend to be much lower than the score threshold. Thus, there are almost no false positives used for the supervision. At the following stage, some fewer false positives, not filtered by the co-generation module, are indeed used for the supervision as mentioned above. However, compared to the dominant unlabeled true positives, the limited false positives cannot effectively interference detector learning. We will exploit to use noisy label methods or uncertainty approaches to further reduce their impact.



\section{Conclusion}

In this paper, we propose a simple but effective mechanism, called \textit{Co-mining}, for the sparsely annotated object detection. \textit{Co-mining} contains a co-generation module and a Siamese network. The co-generation module aims to convert the unlabeled instances as positive supervisions, while the Siamese network aims to generate more diverse pseudo labels using two different branches with the inputs of original images and augmented images. our \textit{Co-mining} is a general method that can be applied to both anchor-free and anchor-based methods. \textit{Co-mining} outperforms other methods on sparsely annotated versions of MS COCO dataset.

\section{Acknowledgements}
This work was supported by The National Key Research and Development Program of China (No. 2020AAA0105200) and Beijing Academy of Artificial Intelligence (BAAI).

\bibliography{egbib}

\end{document}